\documentclass{amia}
\usepackage{graphicx}
\usepackage{bbm}
\usepackage{amsmath,bm}
\usepackage[labelfont=bf]{caption}
\usepackage[style=numeric,sorting=none,sortcites=true,maxnames=6,minnames=3,autocite=superscript]{biblatex}
\usepackage{color}
\usepackage{tabularx}
\usepackage{booktabs}
\usepackage{longtable}
\usepackage{makecell}
\usepackage{multirow}
\usepackage{placeins}
\usepackage{setspace}
\usepackage{amssymb}

\begin{document}

\title{Multi-task Learning via Adaptation to Similar Tasks\\
for Mortality Prediction of Diverse Rare Diseases\footnote{This work
has been submitted for consideration to the AMIA 2020 Annual Symposium}}

\author{Luchen Liu$^{1}\footnote{The two authors have equal contribution.}$,~~ Zequn Liu$^{1*}$,~~ Haoxian Wu$^{1}$,~~ Zichang Wang$^{1}$,\\ Jianhao Shen$^{1}$,~~ 
Yipiing Song$^{1}$,~~ Ming Zhang, PhD$^{1}$}

\institutes{
    $^1$ Department of Computer Science, Peking University, Beijing, China 
}

\maketitle

\noindent{\bf Abstract}

{\it
Mortality prediction of diverse rare diseases using electronic health record (EHR) data 
is a crucial task for intelligent healthcare.
However, data insufficiency and the clinical diversity of rare diseases  make it hard for directly training deep learning models on individual disease data or all the data from different diseases.
%Clinical mortality of patients with rare diseases is highly related to the complex temporal patterns in longitudinal data from electronic health record (EHR). 
Mortality prediction for these patients with different diseases can be viewed as a multi-task learning problem with insufficient data and large task number.
 %Multi-task learning methods can be used to solve the problem of disease behavior diversity by learning task-specific patterns,
But 
the tasks with little training data also make it hard to train task-specific modules in multi-task learning models.
%Meta-learning can deal with data insufficiency by learning the initialization as meta-knowledge for fast adaptation to tasks with very few data, but the important information of task similarity is ignored in these generic meta-learning methods.
To address the challenges of data insufficiency and task diversity, we propose an initialization-sharing multi-task learning method (Ada-Sit) which learns the parameter initialization for fast adaptation to dynamically measured similar tasks.  We use Ada-Sit to train long short-term memory networks (LSTM) based prediction models on longitudinal EHR data. 
 And experimental results demonstrate that the proposed model  is effective for mortality prediction of diverse rare diseases. 
}

\section{Introduction}
Mortality prediction~\cite{Sharma2017Mortality} of diseases plays a  crucial role in clinical work, which could help doctors to take early intervenes based on timely alert of patients' adverse health status.
  With the immense accumulation of Electronic Health Records (EHR) available~\cite{eicu,johnson2016mimic}, deep learning models~\cite{liu2018learning}, typically requiring a large volume of data, have been developed and demonstrate state-of-the-art performance on mortality prediction of common diseases.
  However, mortality prediction of rare diseases is relatively unexplored in the domain of intelligent healthcare and personalized medicine.

Predicting mortality of rare diseases suffers from the problem of data insufficiency.
A rare disease is the disease that affects a small percentage of the population and has a different disease mechanism from common ones.
However, there are more than 300 million people worldwide living with one of the approximately 7,000 rare diseases (US organization Global Genes$\footnote{https://globalgenes.org/rare-list/}$).
Therefore, there is always not enough volume of data for a specific rare disease.
In some real-world data sources~\cite{johnson2016mimic}, only tens of data samples on average could be collected for each rare disease. 

Besides data insufficiency, 
clinical behavior diversity of these diseases is also another challenge for mortality prediction of rare diseases.
Behaviors of different diseases vary a lot and make potential conflicts to the training of global deep learning models~\cite{hochreiter1997long,bai2018empirical,vaswani2017attention}.
For example, the high heart rate raises the mortality risk of people with heart disease, but for patients having a cold, it is a common symptom that does not indicate the patient will be in danger.
So the problem of data volume constraint of diverse rare diseases cannot be simply resolved by training a global model based on  samples from all diseases.
%As we can see, training altogether with samples from diseases with diverse behavior may confuse   the predictive model.

Multi-task learning models~\cite{song2018attend,Suresh2018Learning} can be used to settle the problem of disease behavior diversity. 
Mortality prediction for each kind of rare disease is viewed as a task and  
multi-task learning is supposed to capture the task-specific characteristics as well as the shared information of all tasks.
%For example, the sharedscheme of most neural multi-task learning methods is feature-level sharing, where a subspace of the clinical feature or patient representation is shared across all the tasks~\cite{chen2018meta}. 
However,
the rare disease tasks with little training data make it hard to train specific modules in multi-task learning models,
and further utilizing the shared characteristics of similar tasks is difficult. 

 Meta-learning methods can learn meta-knowledge of training models, which makes it possible to learn fast with few samples, such as few-shot learning~\cite{finn2017model}.
To  build a better multi-task model suitable for  tasks with little training data, 
we bring in the idea of fast adaptation in  meta learning, which  learns a shared initialization  as meta-knowledge for adaptation to new tasks.
%with the ability of measuring task similarity.
However, since the fast adaptation method adapts shared initialization to each task independently, 
%without using the information of similar tasks.
it cannot directly take into account the relationship of similar tasks, 
which  is important and can provide useful information to enhance multi-task learning~\cite{evgeniou2005learning,jacob2009clustered}.
\begin{table*}[htbp]
%\hspace{5cm}
\begin{center}
\begin{tabular}{|c|c|c|c|c|}
\hline
\textbf{Method}&\textbf{Examples}&\multicolumn{3}{c|}{\textbf{Challenges}} \\
\cline{3-5} 
  &   & \textbf{\textit{Data Insufficiency}}& \textbf{\textit{Task Similarity}}& \textbf{\textit{Task Diversity}} \\
\hline
\textbf{Global Models}  & LSTM~\cite{hochreiter1997long}, TCN~\cite{bai2018empirical} & $\checkmark$ & $\checkmark$ & $\times$  \\
\hline
\textbf{Multi-task Models} & Multi-SAnD~\cite{song2018attend}, Multi-Dense~\cite{Suresh2018Learning} & $\times$  & $\checkmark$ & $\checkmark$ \\
\hline
\textbf{Meta-learning Models} & MAML~\cite{finn2017model} & $\checkmark$ & $\times$  & $\checkmark$ \\
\hline
\textbf{Our Model} & Ada-SiT & $\checkmark$ & $\checkmark$ & $\checkmark$ \\
\hline
\end{tabular}
\caption{Comparing the three kinds of methods in terms of their ability to handle three main challenges in mortality prediction for rare diseases}
\label{tab:chal}
\end{center}
\end{table*}

To deal with the above-mentioned challenges (summarised in Table \ref{tab:chal}), we propose an initialization-shared multi-task learning method, named as Ada-SiT (\textbf{Ada}ptation to \textbf{Si}milar \textbf{T}ask), in which the task similarity is dynamically measured in the process of meta-training according to our new definition.
Therefore, the task similarity, as a part of learned meta-knowledge, can enhance the fast adaptation procedure of genetic meta-learning models.
%and samples in similar tasks are gathered to assist the adaptation to each task.
%Specifically, Ada-SiT  learns the  appropriate initialization for fast adaptation to each disease task 
%and dynamically  measures the similarity of all tasks
%in model space to assist the process of task adaptation.
Moreover, Ada-SiT is model-agnostic and can employ all existing deep learning based approaches as the basic predictive model.
%Share Initialization(inspired by gradient-based Meta learning)
%Dynamically measure the task similarity in model space in the training process
%Use samples in Similar tasks to assist the adaptation to a particular task
 Experimental results on real medical datasets demonstrate that the proposed model  is able to make similar tasks cooperate in
 initialization-shared multi-task learning,
 and it outperforms state-of-the-art global models as well as multi-task methods for mortality prediction of diverse rare diseases.
 
 It is worthwhile to highlight the contributions of the proposed model as follow:
 
 \begin{itemize}
    \item To the best of our knowledge, this is the first attempt  to simultaneously tackle the challenge of disease diversity and data insufficiency in mortality prediction of rare diseases. 
    \item We propose a novel initialization-shared multi-task learning method Ada-SiT, which can utilize  information of task similarity for adaptation to each small sample-size task. 
    %Ada-SiT is also suitable for other multi-task learning scenarios with insufficient samples and multiple diverse tasks, besides mortality prediction in the healthcare domain.
    
    %\item  Ada-SiT provides a new task similarity measurement method based on the model parameters of each task, which can be applied to other multi-task learning methods.
    %\item We propose a data preprocessing framework for real-world EHR databases. This framework can transform temporal relational data in the original EHR database into heterogeneous temporal events for both offline training and online prediction, which could have potential benefit for facilitating comparable experimental studies of clinical predictive models based on different EHR data sourses.  
    %The source code will be online available upon acceptance.
\end{itemize}

\section{Related Works}

\subsection{Deep Learning for Healthcare}
The accumulation of Electronic Health Records (EHR) has enabled research on deep learning methods for healthcare~\cite{liu2019learning,wang2019predictive,liu2019early}. Multi-layer Perceptron (MLP)~\cite{cheKDD15}, Convolutional Neural Network (CNN)~\cite{suo2017personalized} and Recurrent Neural Network (RNN)~\cite{choi2016retain,suo2017multi,liu2018learning} have been used in healthcare domain. Among these methods, there are many works on mortality prediction. The good performance of these models depends on a large volume of EHR data, which cannot be satisfied in our scenario of mortality prediction for diverse rare diseases.
%so plenty of patients with different diseases are used to train the model.
As a result, these models cannot make precise mortality predictions for patients with different rare diseases.
Our work is suitable for these settings because it simultaneously tackles the challenges of disease diversity and data insufficiency.
Furthermore, our method is a general framework and can be applied to train deep learning models to improve their performance.

%Transfer learning methods in  healthcare~\cite{purushotham2016variational} aim to learn good clinical prediction models for minority patient groups such as patients with rare diseases.
%They solve the data-insufficiency problem by transferring knowledge from large datasets.
%But our method aims to solve this problem without the help of large datasets by jointly learning multi-task models from multiple datasets with little data.

%Among these methods, there are many works on mortality prediction. Ghassemi~\shortcite{ghassemi2015multivariate} uses a subset of EHR event sequences for prediction while Liu~\shortcite{liu2018learning} models all kinds of events with multi-scale sampling rates.

%The above-mentioned works depend on a large volume data, so plenty of patients with different diseases are used to train the network. As a result, the model cannot capture the diverse disease behaviors and predict mortality for patients with rare diseases. Our work solve this problem by simultaneously tackling the challenge of disease diversity and data insufficiency and can be applied to improve any of these works since it's a general framework.

\subsection{Multi-task Learning}

%In traditional non-neuron models, some methods~\cite{argyriou2007multi,negahban2008joint} use block-sparse regularization to enforcing sparsity across tasks, assuming that all models share a small set of common features, others~\cite{evgeniou2005learning,jacob2009clustered,zhou2011clustered,gong2012robust,ma2018modeling} learn the task relationships, which is related to our work. Our work has different mechanism of task clustering and similarity measuring from these works. For example, Jacob~\cite{jacob2009clustered} and Zhou~\cite{zhou2011clustered} utilize the prior knowledge that tasks are clustered into groups, while our method does not have pre-defined task clusters and finds similar tasks dynamically in the training stage.

Multi-task learning is an efficient method to improve the performance by jointly learning multiple related tasks. 
In deep multi-task learning models, the information sharing mechanism is based on specific network structures, including shared layers~\cite{long2015learning}, shared functions~\cite{chen2018meta} and additional constraints ~\cite{misra2016cross}.
However, task similarity cannot be directly interpreted in these models . We propose a model-agnostic multi-task learning method which can share the parameter initialization for fast adaptation to each task and task similarity can be dynamically measured in the training prossess.

In the healthcare domain, multi-task learning is used for prediction of various clinical events~\cite{harutyunyan2017multitask}, mortality prediction of multiple patient cohorts~\cite{Suresh2018Learning} and patient-specific diagnosis~\cite{nori2017learning}, in which the "tasks" have different definitions. Similar to the paper~\cite{Suresh2018Learning}, our work also treats the mortality prediction of a certain patient cohort as a task.
However, the method proposed by Suresh~\cite{Suresh2018Learning} is suitable for a small number of tasks with a large volume of data, meanwhile, our method on mortality prediction for rare diseases is designed to deal with hundreds of tasks with insufficient data.
%However, the method proposed by Suresh~\cite{Suresh2018Learning} has a small amount of tasks with a large volume of data, while our method has hundreds of tasks with insufficient data due to the feature of rare diseases.

\subsection{Optimization-based Meta Learning}

To solve the problem of data insufficiency and task diversity, our method borrows the idea behind optimization-based meta-learning~\cite{finn2017model,andrychowicz2016learning}, which can adapt to new environments with a few training samples by modifying the parameter optimization process. 
MAML (Model Agnostic Meta Learning)~\cite{finn2017model} uses fast adaptation to find a good initialization for the parameters and adapt it to new tasks.
%MAML (Model Agnostic Meta Learning)~\cite{finn2017model} can %ind the good initialization for the parameters of new tasks.
%by optimizing both the parameter initialization and each task's parameters.
Our work is similar to MAML for using fast adaptation to get parameters of each task, but different from it in two ways: First, the objective of MAML is to learn a good parameter initialization for fast adapting to new tasks but our work is to find good model parameters for each given task. Second, our work measures task similarity in model space dynamically and uses samples in similar tasks to assist the adaptation to each task while MAML does fast adaptations to each task independently.

In clinical scenario, MetaPred~\cite{Zhang2019MetaPred} use MAML for clinical risk prediction with limited patient electronic health records. Its task is similar to ours but its method is different from ours in two ways: First, it trains a parameter initialization on source domain and simulated target domain via fast adaptation, but our method can learn from multiple small target domains without the source domain knowledge. Second, like MAML, MetaGred also doesn't consider task similarity.

\section{Data and Task Descriptions}
We give the notations and data descriptions of the predictive tasks in the following.

\subsection{Heterogeneous Temporal Events in EHR data}
\label{heter}
\label{sec:heterogeneous_temporal_events}
The input
$\bm X$ of each mortality prediction task is a given episode of patient EHR,
 which could be represented as $T$ heterogeneous temporal events~\cite{liu2018learning}:
$\bm X = \{e_t\}_{1\leq t\leq T}$.
$e_t$ in this sequence is a tuple with four element: 
$e=(type, value_c, value_n, time)$
%$e_t = (type,value,time)$
, where $type$ is the clinical event type, $value_c$  and $value_n$ are the categorical and numerical attributes of $e_t$, and $time$ is the record time of $e_t$. 

\subsection{Multi-Task Mortality Prediction}

 The mortality prediction of each rare disease is defined as  a task. 
 Specifically, assuming that there are M diseases, 
  we refer $task_i$ as the
corpus of the i-th mortality prediction task for patients of rare disease $d_i$  with $N_i$ samples:
 \begin{equation}
\setlength{\abovedisplayskip}{3pt}
 task _ { i } = \left\{ \left( \bm X _ { k } ^ { ( i ) } ,\hat{y_ { k } ^ { ( i ) }} \right) \right\} _ { k = 1 } ^ { N _ { i } }
 \setlength{\belowdisplayskip}{3pt}
\end{equation}
where$\bm X _ { k } ^ { ( i ) } $ and $\hat{y_ { k } ^ { ( i ) }}$
denote the k-th sample and its label respectively in the i-th task.

Specifically,
$\bm X$ is a given episode of a
patient's EHR data,
 represented as heterogeneous temporal events, 
 and $\hat{y}$ is the binary label indicating whether a patient will die in 24 hours.

\subsection{Patient Cohort Setup}
\subsubsection{Heterogeneous Temporal Event Datasets}
We set up two heterogeneous temporal event datasets based on MIMIC-III~\cite{johnson2016mimic} database and eICU~\cite{eicu} database. MIMIC-III is a large, freely-available database comprising health data of patients in critical care units from Beth Israel Deaconess Medical Center between 2001 and 2012, and eICU is populated with data from a combination of many critical care units throughout the continental United States between 2014 and 2015. 

The two datasets have the same data preprocessing framework: For each patient, we select all the events with their features from the original database and arrange the events in the temporal order. The descriptions of the selected events in eICU are listed in Table~\ref{tab:eicu_event}. The details of the events in MIMIC-III can be found in \cite{amia}. Then we annotate the mortality label for each patient event sequence. 
\begin{table*}
\centering
\begin{tabular}{lc}  
\toprule
\textbf{Table Name in eICU} & \textbf{Description}\\
\midrule
lab & Laboratory tests mapped to a standard set of measurements. E.g. labTypeID, labResult \\
intakeoutput & Intake and output recorded for patients. E.g. intakeTotal, outputTotal, dialysisTotal \\

medication &
Active medication orders for patients. E.g. drugHiclSeqno, dosage\\
infusiondrug &
Details of drug infusions. E.g. drugRate, infusionRate, drugAmount\\
careplan &
Documentation relating to care planning. E.g. cplGeneralID, cplItemValue\\
admissiondrug &
Details of medications that a patient was taking prior to admission to the ICU. E.g. drugUnit\\
nursecharting &
Information entered in a semi-structured form by the nurse. E.g. nursingchartcelltypecat\\
physicalexam & 
Patients' results of the physical exam. E.g. physicalExamText, physicalExamValue\\
diagnosis &
Diagnosis information of Patients. E.g. ICD9Code, diagnosisPriority\\
respiratorycare &
Information related to respiratory care for patient. E.g. airwayType, airwaySize, cuffPressure\\
allergy &
Details of patient allergies. E.g. allergyType\\
\bottomrule
\end{tabular}
\caption{
The tables in eICU to construct heterogeneous temporal events
%The tables we choose to construct clinical event sequences in MIMIC-III.
}
\label{tab:eicu_event}
\end{table*}

\subsubsection{Rare Disease Selection}
We reorganize the heterogeneous temporal event datasets to get two rare disease datasets, MiniMIMIC and MiniEICU. 

For each ICD (International Classification of Diagnose $\footnote{http://www.icd-code.org}$) code in MIMIC-III and eICU, we calculate its sample size (i.e. the number of patients with this code). We select 858 ICD codes with less than 40 samples in MIMIC-III and 70 ICD codes with less than 100 samples in eICU as rare diseases. The heterogeneous temporal event sequences of patients with selected rare disease $d_i$ form $task_i$. 
%As defined in Section ~\ref{notation}, we construct a heterogeneous event sequence and a binary label meaning death or discharge for each patient in $task_i$ according to his clinical records in MIMIC-III. %Each $task_i$ has less than 40 patients. 
So $MiniMIMIC$ is the task list $\{task_1, ..., task_{858}\}$ and $MiniEICU$ is the task list $\{task_1, ..., task_{70}\}$. 

The statistics of $MiniMIMIC$ and $MiniEICU$ are summarized in Table \ref{tab:statistics}.

Each $task_i$ is split into 3 parts with fixed proportions, namely $Train_i$(70\%), $Valid_i$(10\%) and $Test_i$(20\%). The validation set is used for conducting ``early stop'' and selecting hyper-parameters.
The results of the evaluation metrics on the test set and their stander variations are used to compare different models.

%The decision time sequence is defined as an equal difference progression. 
%The last item is the time 24 hours ahead of the target event, and the difference is 12 hours.

%\subsection*{Clinical endpoint prediction task}

%The clinical endpoint prediction task is formulated as follow: given a clinical heterogeneous event sequence $\{e_t\}_{T_{start}\leq e_t.t\leq T_{end}}$, and a sequence of time stamps $T_{decision}=\{t_i\}$ called decision time sequence in the following, the problem is, for each $t_i \in T_{decision}$, to predict whether the target event will occur afterwards using $\{e_t\}_{T_{start}\leq e_t.t\leq t_i}$.

\section{Methodology}

In this section, we begin with some basic notations for the adaptation to a single task and the backbone deep network for mortality prediction based on EHR data. 
Then we introduce the framework of our method Ada-Sit (Adaptation to similar tasks), which could be applied to train multiple mortality predictive models for different rare disease. 
%Finally, we represent how Ada-Sit is applied to mortality prediction for diverse rare diseases based on heterogeneous temporal events in EHR data.

%\subsection{Notations}
%\label{notation}

%Each event is a triple $e_i = (type,value,time)$, where $type$ is the category of event, $value$ is the attribute of the event, and $time$ is the logged time of $type$ and $value$. 
%$\{e_i\}_{T_{start}\leq e_i.time\leq T_{end}}$ , denoted as $\bm X$, is an input sequence of events $e_i$ in the ascending order of $e_i.time$ in time period $[T_{start},T_{end}]$.

%Given $\bm X$, the clinical outcome prediction task is to dynamically predict
%either death in hospital or discharge to home.

%\subsubsection{Adaptation to a Single Task}
\subsection{Adaptation to a Single Task}
%notation

%single learning
For a task $task_i$ of a given disease and given initial parameters $\theta$ (either random or learned), we formulate the learning process $\operatorname{Learn}(task_i; \theta)$  of model parameters $\theta_i$  as minimizing the loss function of model parameters on the data of the given task from the initialization $\theta$.
\begin{equation}
\setlength{\abovedisplayskip}{3pt}
\theta_i = \operatorname { Learn } \left( task_i ; \theta  \right) = \arg \min _ { \theta } \mathcal { L } _{ task_i } ( \theta )
\setlength{\belowdisplayskip}{3pt}
\end{equation}

We assume $p(y | \bm X,\theta)$ is the mortality rate predicted by model with parameters $\theta$. The loss function $\mathcal { L } _{ task_i } ( \theta )$  of model parameters $\theta$ on task corpus $task_i$ is defined using the cross entropy $\operatorname{CE}(\cdot)$ between the model output $p(y |\bm X,\theta)$ and the true label $\hat{y}$ of the outcome:
\begin{equation}
\setlength{\abovedisplayskip}{3pt}
\mathcal { L } _{ task_i } ( \theta ) = \sum_{(X,\hat{y}) \in task_i} \operatorname{CE}(p(y |\bm X,\theta), \hat{y})
\setlength{\belowdisplayskip}{3pt}
\end{equation}

%Ada-Sit can be applied to mortality prediction tasks on EHR data.
In this work, the mortality prediction model $p(y |\bm X,\theta)$ based on heterogeneous temporal events in EHR data is mainly composed of attributed event embedding~\cite{liu2018learning} and long short-term memory (LSTM). Firstly, clincial events with attributes are embedded into vectors via information fusion of their type,  categorical attributes and numerical attributes. After the attributed event embedding module, the temporal dependencies encoded in the sequence of embedded vectors are then captured by long short-term memory (LSTM) \cite{hochreiter1997long}, which outputs the prediction results for mortality with a sigmoid layer at the last LSTM cell. The mortality prediction model is illustrated in the middle of Figur \ref{fig:motality prediction model}.

\subsection{Adaptation to Similar Tasks}
\begin{figure*}[!t]
\centering
\includegraphics[width=.99\textwidth]{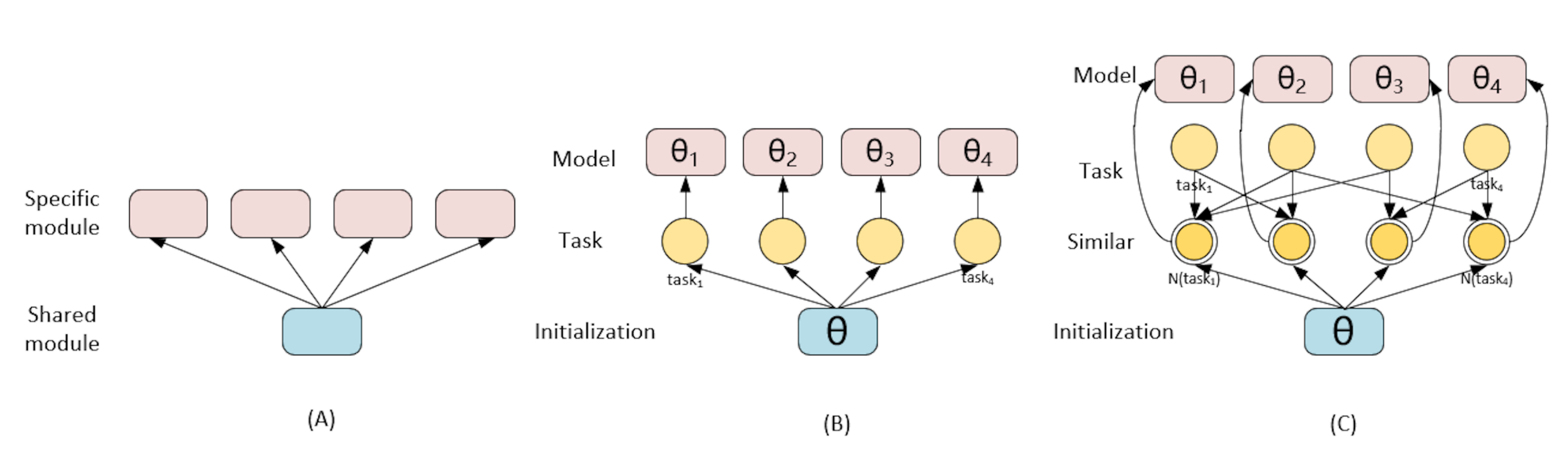}
\caption{(A) Generic Architecture of Multi-Task Learning. The
shared module (in blue), whose output will be taken as the input of specific modules,
is shared between different tasks.
(B) Generic Architecture of Meta-Learning based Multi-Task Learning. The parameter initialization $\theta$ in blue, which will be adapted to each specific task $task_i$ , is shared between all tasks
(C) Architecture of our Ada-SiT (\textbf{Ada}ptation to \textbf{Si}milar \textbf{T}ask). The shared initialization $\theta$ will be adapted to similar tasks $N(task_i)$ of each task $task_i$, resulting in a predictive model $\theta_i$ for each task.}
\label{fig:model}
\end{figure*}

\begin{figure*}[!t]
\centering
\includegraphics[width=.95\textwidth]{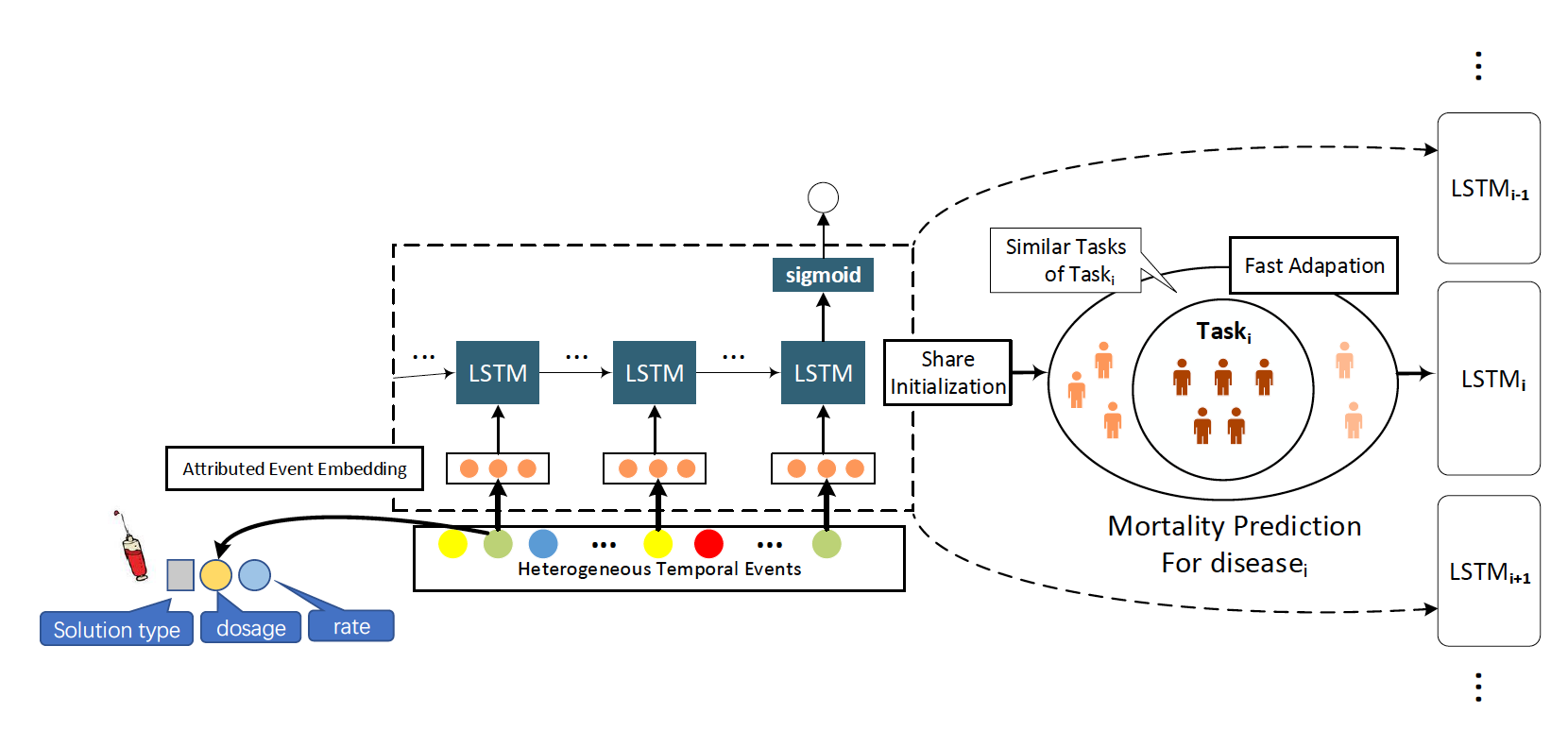}
\caption{Training Mortality Prediction Model with Ada-Sit}
\label{fig:motality prediction model}
\end{figure*}

The architecture of our proposed multi-task learning method Ada-SiT is represented in Figure \ref{fig:model}, where Ada-Sit is compared with generic multi-task methods and  generic meta-learning based multi-task methods.
In the following,
we first introduce the architecture of Ada-SiT,
%which is suitable for multi-task learning with insufficient data in each task.
 %Ada-Sit is also able to utilize the relationship information of similar tasks for
 and then the task similarity measurement module, which is a key component of Ada-SiT.
The overall framework of how to train mortality prediction model with Ada-Sit is illustrated in Figure \ref{fig:motality prediction model}.

\subsubsection{Architecture of Ada-SiT
}\label{sec:ada-sit}

%To build connections among all the small-size tasks,
Fast adaptation,
one of the meta-learning methods, is applied into Ada-SiT for the
multi-task learning scenario.
Ada-SiT differs from the original fast adaptation ~\cite{finn2017model} which separately adapts to each task.
Ada-SiT can dynamically measure task similarity and learn to adapt to multiple similar tasks.
%and explore their relationship of   the similar tasks.

%MAML idea
The idea of fast adaptation is to learn a good parameter initialization for fast adaptation to new tasks.
Specifically, in our scenario of multi-task learning,
it means  using patient data from all the disease tasks to find a good initial parameter initialization $\theta$ and  mortality prediction model parameters $\theta_i$ of each  disease which is adapted from the found initial parameters $\theta$.

%objective function
The initialization $\theta$  is learned by repeatedly simulating scenarios of  mortality prediction of each disease with its similar diseases.
We achieve this goal by defining the meta-objective function as:
\begin{equation}
\setlength{\abovedisplayskip}{3pt}
 \min _\theta 
 %\sum_{task_i} \mathcal { L }_{N(task_i)}  \left(  \theta'_i  \right) \\
 %=
 \sum_{task_i} \mathcal { L }_{N(task_i)}  \left(  \operatorname{Learn}(N(task_i);\theta)  \right)
\setlength{\belowdisplayskip}{3pt}
\end{equation}
where $N(task_i)$ is the extended sample set composed of tasks similar with $task_i$.
Details of $N(task_i)$ will be described in the next subsection.

%single task update
We maximize the meta-objective function using stochastic approximation with gradient descent.
For each epoch, we find similar tasks $N(task_i)$ for each task, 
and then independently sample two samples subsets ($D_{tr}$ and $D_{val}$) from the training set of the task $N(task_i)$.
The former $D_{tr}$ is used to simulate the learning process of mortality prediction models, and the latter $D_{val}$ is used to evaluate the precision of the learned models for the updating of the shared initialization.
Here, a single-step gradient descend is applied in the training simulation:
\begin{equation}
\setlength{\abovedisplayskip}{3pt}
\theta _ { i } ^ { \prime} = \operatorname { Learn } \left( N(task_i  ) ; \theta \right) = \theta - \alpha \nabla _ { \theta } \mathcal { L } ^{ tr} _ {N(task_i )} ( \theta ) 
\setlength{\belowdisplayskip}{3pt}
\end{equation}
where $\alpha$ is the learning rate and $ \mathcal { L } ^{ tr} _ {N(task_i )} ( \cdot  )$ represents the loss function calculated from sample set $D_{tr}$ from $N(task_i)$.

%initialization update
Next, we evaluate the updated task parameters $\theta_i'$ on $D_{val}$.
The gradient computed from the evaluation, referred to as meta-gradient, is used to update the initial parameters $\theta$.
Gradients from a batch of tasks are aggregated to updating   $\theta$ as follow:
\begin{equation}
\setlength{\abovedisplayskip}{3pt}
\theta \gets \theta - \beta \nabla _ { \theta } \sum_{task_i} \mathcal { L }^{val}_{N(task_i)}  \left(  \theta'_i  \right) 
\setlength{\belowdisplayskip}{3pt}
\end{equation}
where $\beta$ is the meta learning rate,
and $ \mathcal { L } ^{ val} _ {N(task_i )} ( \cdot  )$ represents the loss function calculated from sample set $D_{val}$ from $N(task_i)$.

%simplified gradient
In the process of the calculation and approximation of the meta-gradient by the chain rule, the second-order derivative term can be ignored without much accuracy loss~\cite{finn2017model}. And the meta-gradient can be approximated as the following simplified gradient:
\begin{equation}
\setlength{\belowdisplayskip}{3pt}
\begin{array} {  l}{\nabla _ { \theta } \mathcal { L }^{val} _{ N(task_i) } \left( \theta ^ { \prime } \right) = \nabla _ { \theta ^ { \prime } } \mathcal { L } ^{val}_{ N(task_i)  } \left( \theta ^ { \prime } \right)  \left( 1- \alpha H _ { \theta } \big( \mathcal { L }^{tr} _{ N(task_i) } ( \theta ) \big) \right) \approx \nabla _ { \theta ^ { \prime } } \mathcal { L } ^{val}_{ N(task_i)  } (\theta^ { \prime })}
\end{array}
\setlength{\belowdisplayskip}{3pt}
\end{equation}
where the term including $H_ \theta(\mathcal{L}^{tr}_{N(task_i)}( \theta ) )$, the Hessian matrix, a square matrix of second-order partial derivatives of the loss function $\mathcal{L}^{tr}_{N(task_i)}( \cdot )$ at $\theta$, is ignored. 
%This item containing the second-order partial derivatives is ignored in practice when we are calculating the meta-gradient.

%each task updating
As our goal is multi-task learning via adaptation to similar tasks, model parameters $\theta_i$ of each task
are adapted from the newly updated initialization parameters $\theta$ at the end of each iteration.
These parameters $\theta_i$ of tasks are comparable in the model space, and will be used for calculating the task similarity in model space in the next iteration.
%Notice that model parameters $\theta_i$ are adapted from the identical  initialization parameter $\theta$ to make them comparable in the model space.

%similarity metric
\subsubsection{Task Similarity Measurement in Model Space}\label{sec:simtask}
To measure  the similarity of all tasks in terms of clinical behavior, we define the similarity of tasks in the model space, where each predictive model for its corresponding task is represented as a vector composed of all the parameters.

Formally, the similar tasks $N(task_i)$ of $task_i$ is defined as $N_{cos}(\cdot)$:
%The models of diseases with similar clinical behaviors tend to give relative consistent prediction results, and have similar parameters. So the Euclidean distance in the model space between similar models is small.
\begin{equation}\label{eq:simcos}
\setlength{\abovedisplayskip}{3pt}
 N_{cos}(task_i) = \{s | s \in task_j \And \cos ( \theta_i - \theta, \theta_j - \theta)  > \eta \} 
\setlength{\belowdisplayskip}{3pt}
\end{equation}
where $\theta_i$ and $\theta_j$ are the parameters of corresponding model of $task_i$ and $task_j$, $\theta$ is the initial parameters. $ \cos( \theta_i - \theta, \theta_j - \theta) $ reflects $\cos$ function of the angle between gradient directions of $task_i$ and $task_j$, and $\eta$ is threshold of the $\cos$ function.
%deviation angle scale of $task_i$.

The models of similar diseases also have similar gradient directions when they are adapted from the initialization.
And the big $\cos$ (close to 1)  value of  included angle  between two gradient directions indicates the tasks are similar.

Notice that a natural alternative way to get similar tasks is selecting k nearest neighbors in the model space.
However, the absolute distance of models is more meaningful than relative distance, because
 the distance of models is generated by the gradient descent of the adaptation process from a common initialization.
So selecting models in the neighborhood of a certain model as its similar models are more suitable in our Ada-SiT method, which is demonstrated by the experimental results in Section \ref{similarity}.

\section{Results and Discussions}
\subsection{Comparing Methods and Experimental Settings}
We compare Ada-SiT to both \textit{global single-task learning }methods and \textit{multi-task learning} methods for mortality prediction. The data size of each task is too small to train separate single-task models, so these baselines have not been included.

The \textbf{\textit{global single-task learning}} models are trained by all the patients in the training set.

$\bullet$ \textbf{LSTM} 
 LSTM~\cite{hochreiter1997long} is used to learn the representation of the heterogeneous event sequence for each patient.  Binary predictions for mortality based on the learned representations can be generated with a logistic regression layer.

$\bullet$ \textbf{TCN}
The architect for prediction is the same as \textbf{LSTM}, except that TCN~\cite{bai2018empirical} is used to learn patient representation vectors from the heterogeneous event sequences instead of LSTM.

The \textbf{\textit{multi-task learning}} methods use $Train_i$ to train the model for each $task_i$ and get prediction results such as predicted label and probability on $Test_i$.

$\bullet$ \textbf{Multi-SAnD}
Multi-SAnD~\cite{song2018attend} is a Transformer-based multi-task learning method which uses the weighted sum of loss functions on all tasks for the loss function.

$\bullet$ \textbf{MMoE}
MMoE~\cite{ma2018modeling} is a Multi-gate Mixture-of-Experts model which shares the expert submodel across all tasks and has a gating network trained to optimize each task. 

%The following two methods are proposed by Suresh~\cite{Suresh2018Learning}. Each model has an LSTM layer for representation learning, followed by a dense layer and a logistic regression output layer.

%$\bullet$ \textbf{Shared LSTM + Separate Output Layers}
%The models for different tasks share parameters in the LSTM layers and dense layers while having task-specific parameters in the output layers. We use \textbf{Multi-Output} for short in the following sections.
%$\bullet$ \textbf{Multi-Output}
%The models for different tasks share parameters in the LSTM layers and dense layers while having task-specific parameters in the outMulput layers. 

%$\bullet$ \textbf{Shared LSTM + Separate Dense and Output Layers}
%The models for different tasks have sharing parameters in the LSTM layers and task-specific parameters in the dense layers and output layers. We use \textbf{Multi-Dense} for short in the following sections.
$\bullet$ \textbf{Multi-Dense}
Multi-Dense is proposed by Suresh~\cite{Suresh2018Learning}. It has a shared LSTM layer for representation learning, followed by task-specific dense layers and logistic regression output layers.
%The models for different tasks have shared parameters in the LSTM layers and task-specific parameters in both dense layers and output layers. 

The models in this section are implemented with Tensorflow~\cite{abadi2016tensorflow} and trained with Adam. 
%We use event type embedding and attribute encoding proposed in Section \ref{lstm} in all the competing methods. 
In Ada-SiT, we set $N(task_i) = N_{cos}(task_i)$, where the $\cos$ function threshold $\eta$ is 0.7, for the task similarity measurement. $\alpha$ and $\beta$ are 0.0005 and 0.001 respectively.

\subsection{Evaluation Metrics}
The data for target prediction tasks are imbalanced labeled. So metrics for binary labels such as accuracy are not suitable for measuring the performance. Similar to the work~\cite{choi2016retain}, we adopt \textbf{AUC} (the area under ROC curves (Receiver Operating Characteristic curves)) and \textbf{AP} (the area under PRC (Precision-Recall curves)) for evaluation. They both reflect the overall quality of predicted scores at each decision time.

\begin{minipage}[htb]{\textwidth}
    \begin{minipage}[b]{0.48\textwidth}
  %\centering
    \begin{tabular}{lrr}  
\toprule
Name & MiniMIMIC & MiniEICU\\
\midrule
\# of tasks & 858 & 70     \\
\# of samples & 16610 & 7000      \\
positive sample rate (mortality rate)  & 7\%  &  13\%   \\
max \# of samples per task & 40  & 100    \\
min \# of samples per task & 10  & 100    \\
mean \# of samples per task & 19.36 & 100\\
\bottomrule
\end{tabular}
\makeatletter\def\@captype{table}\makeatother\caption{Statistics of Datasets}
\label{tab:statistics}
  \end{minipage}
    \begin{minipage}[b]{0.68\textwidth}
   \centering
    \includegraphics[width=0.5\textwidth]{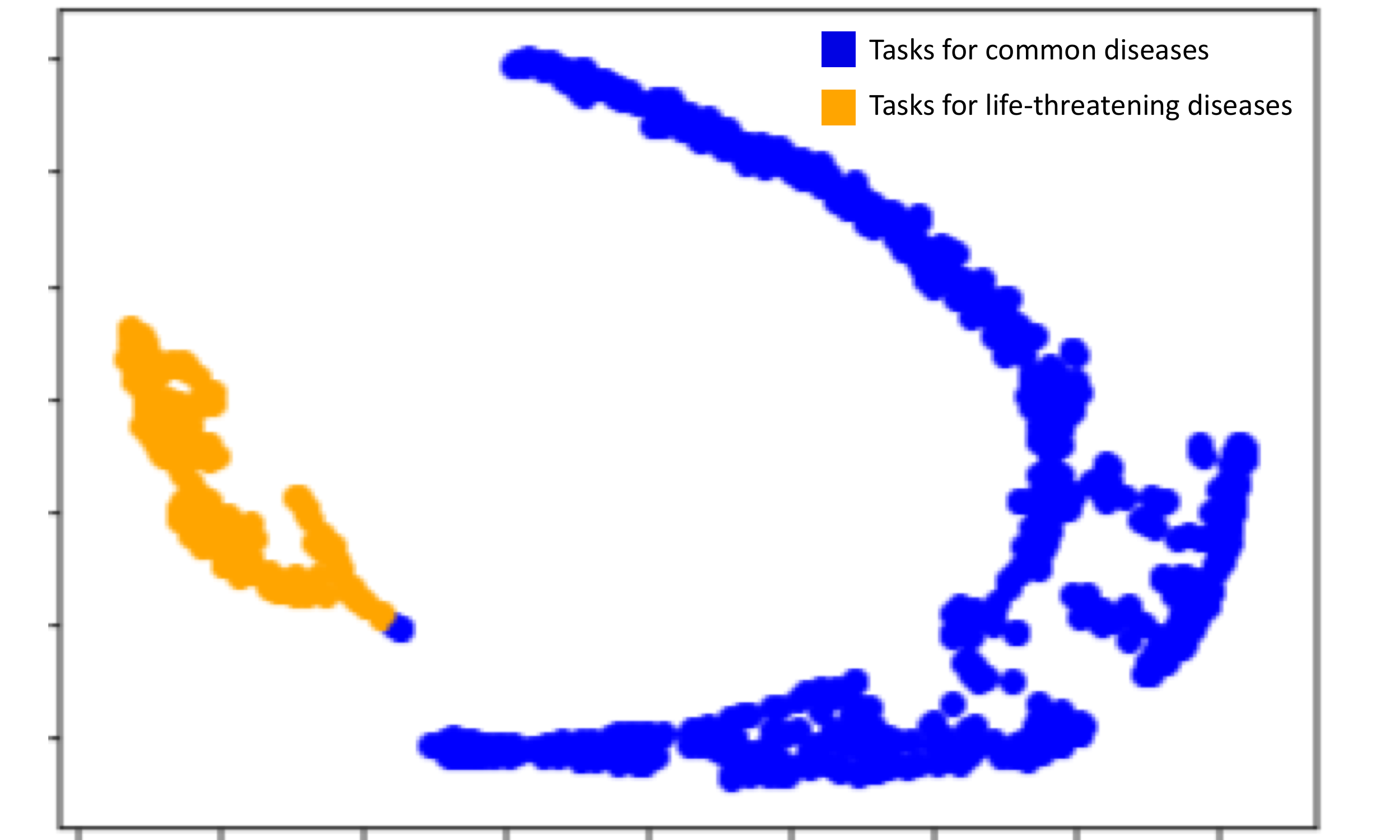}
    \makeatletter\def\@captype{figure}\makeatother\caption{Visualization of Tasks in the Model Space}
    \label{fig:visual}
   \end{minipage}
\end{minipage}
\subsection{Quantitative Results}

\begin{table*}[t]
\renewcommand\arraystretch{1}
\centering
\newcolumntype{P}{>{\centering\arraybackslash}p{2.5cm}}
\begin{tabular}{ll@{\hskip 0.6cm}PP@{\hskip 0.3cm}PP}
\toprule
\multirow{2}{*}{Model Class} &  \multirow{2}{*}{Model} & \multicolumn{2}{c}{MiniMIMIC} & \multicolumn{2}{c}{MiniEICU}  \\
 & & \textbf{AUC} & \textbf{AP} & \textbf{AUC} & \textbf{AP} \\
\midrule
\multirow{2}{*}{Global Single-task} & LSTM~\cite{hochreiter1997long} & 0.8162 (0.0026) & 0.3830 (0.0055) & 0.6642 (0.0227) & 0.2692 (0.0193) \\
 & TCN~\cite{bai2018empirical} & 0.8008 (0.0024) &  0.4120 (0.0011) & 0.6107 (0.0055) & 0.1945 (0.0052) \\
\midrule
\multirow{3}{*}{Multi-task} & Multi-SAnD~\cite{song2018attend} & 0.8036 (0.0161) &  0.2754 (0.0063) & 0.6215	(0.0075) &	0.1592 (0.0016)
  \\
 & MMoE~\cite{ma2018modeling} & 0.7181 (0.0117) & 0.2195 (0.0097) & 0.6300 (0.0023) & 0.1364 (0.0030)
 \\
%\hline
% & Multi-Output & 0.8056 (0.0081) & 0.3449 (0.0091) & \textbf{0.6969 (0.0036)} &	0.0928 (0.0085)\\
%\hline
& Multi-Dense~\cite{Suresh2018Learning} & 0.8325 (0.0036) & 0.3997 (0.0096) & 0.6730 (0.0071) &	0.1147 (0.0039)\\
\midrule
Ours & Ada-SiT & \textbf{0.8729 (0.0112)} &  \textbf{0.4543 (0.0241)} &\textbf{0.6746 (0.0090)} & \textbf{0.2961 (0.0103)}\\
\bottomrule
\end{tabular}
\caption{performance of different models on $MiniMIMIC$ and $MiniEICU$}
\label{tab:performance}
\end{table*}

Table \ref{tab:performance} shows the AUC and AP of Ada-SiT, global single-task models, and multi-task models on $MiniMIMIC$ and $MiniEICU$. From the results in Table \ref{tab:performance}, we draw the following conclusions:

First, Ada-SiT can significantly improve the performance of global single-task learning methods. On both datasets, Ada-SiT performs better than LSTM and TCN. For example, on $MiniMIMIC$, Ada-SiT improves AUC and AP by around 6.9\% and 19.1\% respectively compared to LSTM. We can conclude that Ada-SiT can capture specific characteristics of diverse tasks without being interfered by data conflicts.
%efficiently predict the mortality of patients with diverse diseases by fully utilizing the similarity among tasks.

Second, Ada-SiT outperforms the compared multi-task learning methods. For example, on $MiniMIMIC$, Ada-SiT improves AUC and AP of Multi-SAnD by 8.6\% and 65.0\% respectively. On $MiniEICU$, it improves the AUC of Multi-SAnD and MMoE by 8.5\% and 7.1\% respectively. It should be noted that most of the multi-task baselines do not perform better than the global single-task baselines. 
The possible cause is that task-specific parameters of multi-task models cannot be trained well because of data insufficiency of each task. 
We can conclude that Ada-SiT has a more robust information-sharing mechanism among tasks on small-size dataset compared to the traditional multi-task learning baselines.

\subsection{Ablation Experiments of Task Similarity Measurement}
\label{similarity}
To evaluate the effect of task similarity measuring in Ada-SiT, we vary this module while remaining other parts of the model identified in this section. We implement the following variants of Ada-SiT:

$\bullet$ \textbf{Ada-SiT$^-$}
Ada-SiT$^-$ is Ada-SiT without similar task measurement (i.e. $N(task_i) = task_i$), nearly the same as MAML~\cite{finn2017model}

$\bullet$ \textbf{Ada-SiT (Static)}
According to the work ~\cite{ruder2017learning}, many static features can be used to measure task similarity. In Ada-SiT (Static), we choose the mortality rate as the static feature in the clinical scenarios and use it to measure task similarity instead of the proposed similarity measurement.

%$\bullet$ \textbf{Ada-SiT (Euclidean)}
%We set $N(task_i) = N_{\delta}(task_i)$ instead of $N_{cos}(task_i)$.

$\bullet$ \textbf{Ada-SiT (KNN)}
Ada-SiT (KNN) selects k nearest neighbors instead of neighbors within a certain distance for $N_{cos}(task_i)$ while finding similar tasks.

\begin{table}
\centering
\begin{tabular}{lrr}  
\toprule
Methods & \textbf{AUC} & \textbf{AP} \\
\midrule
Ada-SiT$^-$ (MAML) & 0.8577 (0.0015) & 0.3936 (0.0025)    \\
%\hline
Ada-SiT (Static) & 0.8474 (0.0123) & 0.4143 (0.0144)\\
%\hline
%Ada-SiT (Euclidean) & 0.8657 (0.0118) & 0.4227 (0.0007) \\
%\hline
Ada-SiT (KNN) & 0.8264 (0.0110) & 0.4059 (0.0112) \\
%\hline
Ada-SiT & \textbf{0.8729 (0.0112)} &  \textbf{0.4543 (0.0241)}\\
\bottomrule
\end{tabular}
\caption{Ablation study of task similarity measurement}
\label{tab:MAML}
\end{table}

%It's easy to find that the algorithm of Ada is . 
Table \ref{tab:MAML} shows the results of the ablation experiments of task similarity measurement. We can draw the following conclusions:
First, the information in similar tasks can improve the performance of fast adaptation. Ada-SiT (KNN) and Ada-SiT improve the AP of Ada-SiT$^-$. Ada-SiT also improves the AUC of Ada-SiT$^-$.
Second, the task similarity measurement in model space outperforms the compared static measurement. Ada-SiT improves the AUC and AP of Ada-SiT (Static) by 3.0\% and 9.7\% respectively. It is because traditional task similarity measurements via static features only leverage the metadata of tasks, but our measurement in model space can find potential information from the mapping of samples and labels.
%our task similarity measurement can find the potential similarity among tasks and leverage more information than the predefined features.
Third, finding neighbors within a certain distance as $N_{cos}(task_i)$ is the most suitable way to get similar tasks. For example, Ada-SiT improves the AUC of Ada-SiT (KNN) by around 5.6\%. It is noteworthy that the AUC of Ada-SiT (KNN) is even lower than Ada-SiT$^-$. It is possibly because some tasks in the k nearest neighbors of $task_i$ may be far from $task_i$ in model space and they interfere with the fast adaptation process of $task_i$.

\subsection{Relationship between Task Similarity and Mortality Rate}
\label{death rate}
%With the help of a proper task similarity threshold, we can gather similar tasks and find the relationship between task similarity and mortality rate.
%With the help of a proper task similarity threshold,
There is strong correlation between task similarity and mortality rate.
By treating each model's parameter vector $\theta_i$ as a point in the model space, we use t-SNE~\cite{maaten2008visualizing} to visualize similar task clusters in Figure \ref{fig:visual}. 
In Figure \ref{fig:visual}, the two task clusters represent two types of rare diseases. 
%The blue cluster has tasks whose average mortality rate is 0.6\% and the yellow cluster has tasks whose average mortality rate is 32.1\%. (The average mortality rate of $MiniMIMIC$ is 7\%).
The average mortality rate of diseases in the blue cluster is 0.6\% and that in the yellow cluster is 32.1\%. Meanwhile, the total average mortality rate of $MiniMIMIC$ is 7\%.
%\subsubsection{Visualization of Tasks in Model Space}
%\label{visual}
%By treating parameters $\theta_i$ as a point in model space, we use t-SNE~\cite{maaten2008visualizing} to visualize the parameters $\theta_1$, $\theta_2$, ..., $\theta_{858}$. We use blue for the cluster with low mortality rate and yellow for the cluster with high mortality rate in Section \ref{death rate}. Figure \ref{fig:visual} shows the visualization results. 
%We can see that tasks with different mortality rates are separated quite well.
We can see that the mortality rate is the main factor to determine task similarity.
%It suggests that our task similarity measurement module is reasonable, and diseases with similar mortality rate are similar, which is consistent with the clinical knowledge.
It suggests that our task similarity measurement module is reasonable and consistent with the clinical knowledge because diseases with a lower mortality rate and life-threatening diseases with a  high mortality rate have different clinical behavior.

\section{Conclusion}
In this paper, we propose a novel method Ada-SiT for learning predictive models for diverse tasks with insufficient data.
Ada-SiT has a new task similarity measurement method and a new knowledge-sharing schema, where the shared initialization is learned to fast adapt to similar tasks.
Experiment results show that our method is suitable for mortality prediction of diverse rare diseases, 
and can improve the performance compared to global single-task models and genetic multi-task models.

\begingroup
\setstretch{0.7}
\defbibheading{bibliography}[References]{\section*{\centering #1}}
\printbibliography
\endgroup

% \bibliographystyle{unsrt}
% \begin{thebibliography}{1}
% \setlength\itemsep{-0.1em}
% \bibitem{ref1}
% Pryor TA, Gardner RM, Clayton RD, Warner HR. The HELP system. J Med Sys. 1983;7:87-101.
% \bibitem{ref2}
% Gardner RM, Golubjatnikov OK, Laub RM, Jacobson JT, Evans RS. Computer-critiqued blood ordering using the HELP system. Comput Biomed Res 1990;23:514-28.
% \end{thebibliography}
\end{document}